\documentclass[10pt,twocolumn,letterpaper]{article}

\usepackage{cvpr}
\usepackage{times}
\usepackage{epsfig}
\usepackage{graphicx}
\usepackage{amsmath}
\usepackage{amssymb}
\usepackage{mathtools}
\usepackage{times}
\usepackage{epsfig}
\usepackage{graphicx}
\usepackage{comment}
\usepackage{amsmath,amssymb} 
\usepackage[table]{xcolor}
\usepackage{adjustbox}
\usepackage{color}
\usepackage{units}
\usepackage{amsmath}
\usepackage{amssymb}
\usepackage{xcolor}
\usepackage{subcaption}
\usepackage{dblfloatfix}
\usepackage{setspace}
\usepackage{bbm}
\usepackage{gensymb}
\usepackage{booktabs}
\usepackage{cite}
\usepackage[ruled,lined,linesnumbered]{algorithm2e}

\cvprfinalcopy 

\def\cvprPaperID{****} 

\usepackage[pagebackref=true,breaklinks=true,letterpaper=true,colorlinks,bookmarks=false]{hyperref}
\hypersetup{citecolor=[RGB]{119,185,0}}

\begin{document}
\title{FB-OCC: 3D Occupancy Prediction based on Forward-Backward\\View Transformation
}

\author{
\text{Zhiqi Li}$^{1,2}$,
\text{Zhiding Yu}$^{1}$,
\text{David Austin}$^{1}$,
\text{Mingsheng Fang}$^{2}$,
\text{Shiyi Lan}$^{1}$,
\text{Jan Kautz}$^{1}$,
\text{Jose M. Alvarez}$^{1}$
\\ [0.15cm]
$^1$NVIDIA~~~~$^2$Nanjing University
}
\maketitle
\newcommand\blfootnote[1]{%
  \begingroup
  \renewcommand\thefootnote{}\footnote{#1}%
  \addtocounter{footnote}{-1}%
  \endgroup
}


\begin{abstract}
This technical report summarizes the winning solution for the 3D Occupancy Prediction Challenge, which is held in conjunction with the CVPR 2023 Workshop on End-to-End Autonomous Driving and CVPR 23 Workshop on Vision-Centric Autonomous Driving Workshop. Our proposed solution FB-OCC builds upon FB-BEV, a cutting-edge camera-based bird's-eye view perception design using forward-backward projection. On top of FB-BEV, we further study novel designs and optimization tailored to the 3D occupancy prediction task, including joint depth-semantic pre-training, joint voxel-BEV representation, model scaling up, and effective post-processing strategies. These designs and optimization result in a state-of-the-art mIoU score of 54.19\% on the nuScenes dataset, ranking the 1st place in the challenge track. Code and models will be released at: \url{https://github.com/NVlabs/FB-BEV}.
\vspace{-5mm}
\end{abstract}

\section{Introduction}
3D occupancy prediction, which refers to predicting the occupancy status and semantic class of every voxel in a 3D voxel space, is an important task in autonomous vehicle (AV) perception. Predicting 3D occupancy is important to the development of safe and robust self-driving systems by providing rich information to the planning stack~\cite{hu2023planning}. The challenge track requires participants to developing occupancy prediction algorithms that solely utilize camera input during inference. In addition, the challenge permits the use of open-source datasets and models, which facilitates the exploration of data-driven algorithms and large-scale models. The impact of this challenge is significant by providing a playground for the latest state-of-the-art 3D occupancy prediction algorithms in real-world scenarios.

In the context of challenge, besides our efforts in model structure design, we emphasize the importance of both model scale and model pre-training techniques. This focus stems from several motivations. First, there have been a number of bird's-eye view (BEV) perception solutions with state-of-the-art performance~\cite{li2022bevformer,huang2021bevdet,li2023bevdepth}. These solutions can be adapted to 3D occupancy prediction with certain modifications. However, there is still limited knowledge regarding the impact of large-scale models and pre-training on the occupancy prediction task. As will be reported in this work, the use of large-scale models and pre-training techniques stands as crucial factors contributing to our success.

\section{Method}

In this section, we will present our solution in details with the following aspects covered. Section \ref{MSD} will elaborate on our model design. Section \ref{SP} will discuss the efforts in model pre-training and scaling up. Finally, Section \ref{PP} will outline our post-processing strategies.

\subsection{Model design}\label{MSD}

Here, we give an introduction to FB-BEV. It is known that view transformation is a central module of camera-based 3D perception. In the literature, this module is based on two dominant view transformation strategies: forward projection (represented by List-Splat-Shoot~\cite{philion2020lift}) and backward projection (represented by BEVFormer~\cite{li2022bevformer}). FB-BEV adopts a unified design that leverages both methods, promoting the benefits from each method with improved perception results while overcoming their limitations. In the case of FB-OCC, we use forward projection to generate the initial 3D voxel representation. We then condense the 3D voxel representations into a flattened BEV feature map. The BEV feature map is treated as queries within the BEV space and attends the image encoder features to acquire dense geometry information. The fusion features of the 3D voxel representation and the optimized BEV representations are then fed into the subsequent task head.

In the forward projection module, we adhere to the principles of Lift-Splat-Shoot (LSS)~\cite{philion2020lift} to account for the uncertainty in the depth estimation of each pixel. This allows us to project the image features into the 3D space based on their corresponding depth values. In contrast to LSS, which models BEV features, we directly model 3D voxel representations to capture more detailed information in the 3D space. Additionally, we adopt BEVDepth~\cite{li2023bevdepth} to utilize point clouds in generating accurate depth ground truth, which helps supervise the depth prediction of our model for improved accuracy. LSS tends to produce relatively sparse 3D representations. To tackle this issue, we incorporate a backward projection method to optimize these sparse 3D representations. Considering the computational burden, we employ BEV representation instead of 3D voxel representation at this stage. The backward projection method draws inspiration from BEVFormer~\cite{li2022bevformer}. However, unlike BEVFormer, which employs randomly initialized parameters as BEV queries, we compress the obtained 3D voxel representation into a BEV representation, thereby incorporating stronger semantic priors. Furthermore, our backward projection method leverages the depth distribution during the projection phase, enabling more precise modeling of projection relationships.

\begin{figure*}[htb]
\centering
\includegraphics[width=0.95\textwidth]{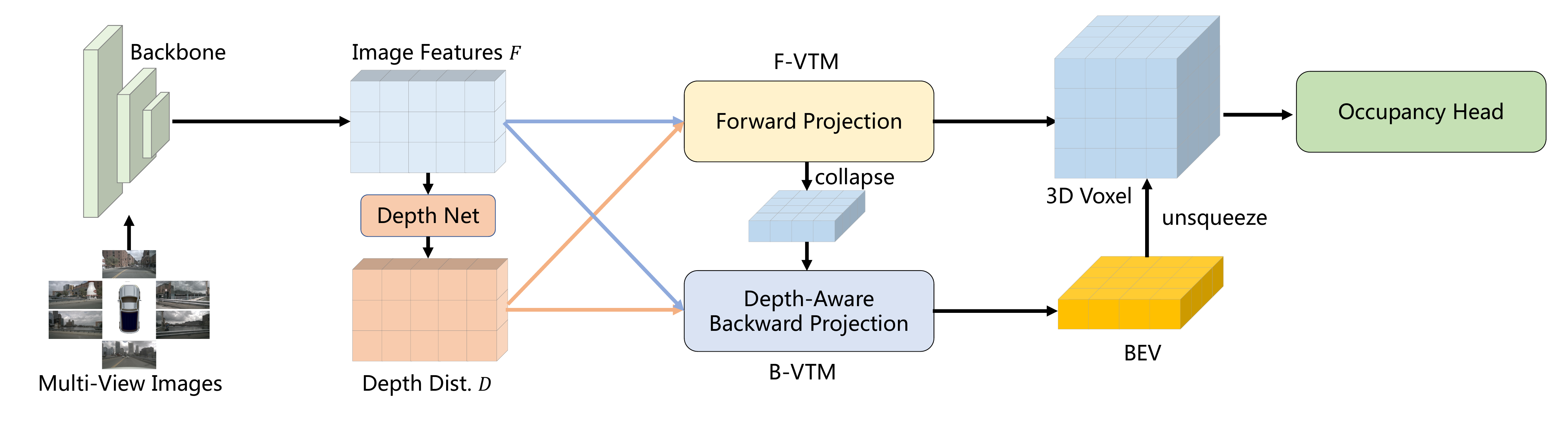}
\vspace{-3mm}
\caption{\textbf{The overall architecture of FB-OCC.} F-VTM is based on forward projection (LSS), and B-VTM is based on backward projection (BEVFormer).
}
\label{fig:model}
\vspace{-3mm}
\end{figure*}

\begin{figure}[t]
\centering
\includegraphics[width=0.95\linewidth]{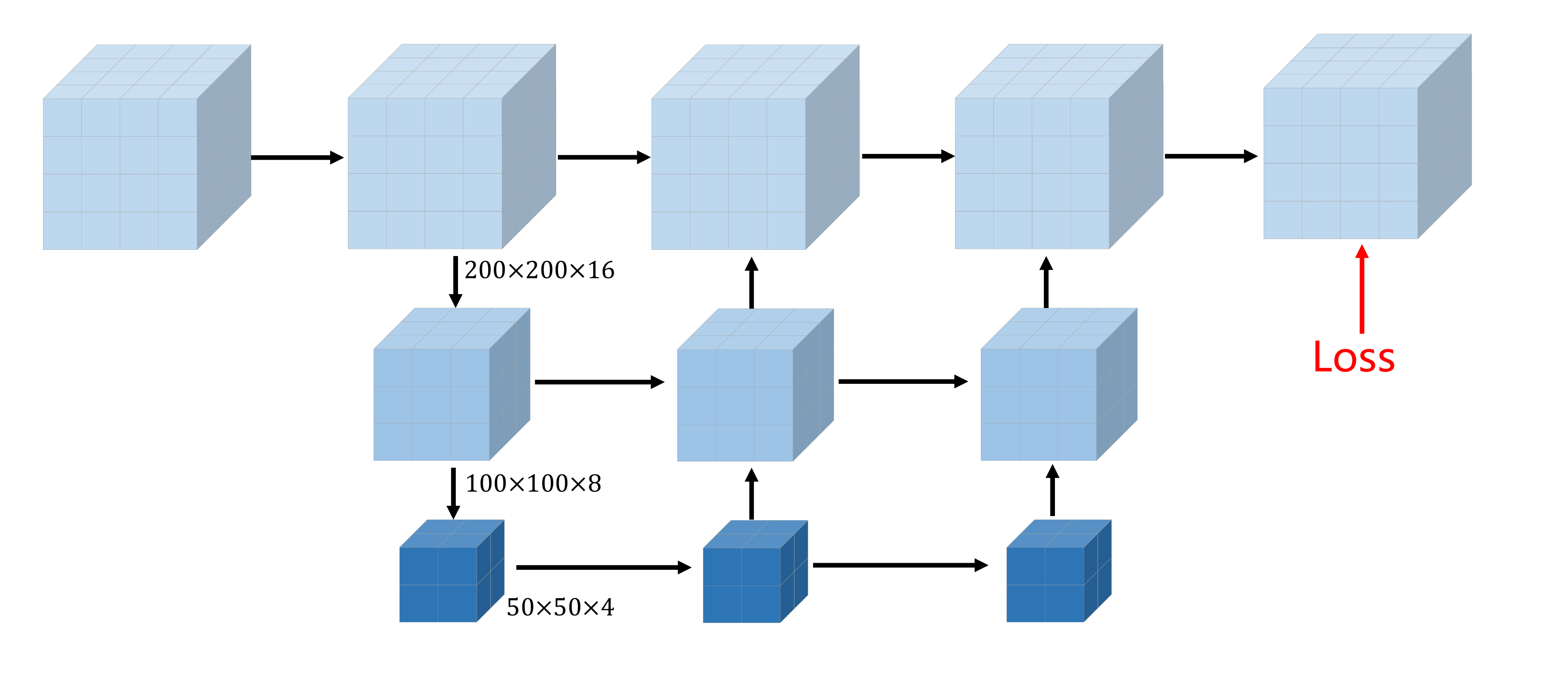}
\vspace{-2mm}
\caption{\textbf{The architecture of the occupancy prediction head of FB-OCC.} 
}
\label{fig:head}
\vspace{-3mm}
\end{figure}

Following the acquisition of the 3D voxel representations and optimized BEV representation, we combine them through the process of expanding the BEV features, resulting in the final 3D voxel representations. The voxel encoder and the occupancy prediction head, as depicted in Figure~\ref{fig:model} and Figure~\ref{fig:head}, are outlined below.

To train the model, we use a distance-aware Focal loss function $L_{fl}$ inspired by M2BEV~\cite{xie2022m2bev}, Dice loss $L_{dl}$, affinity loss
$L_{scal}^{geo}$ and $L_{scal}^{sem}$ from MonoScene~\cite{cao2022monoscene}, lovasz-softmax loss $L_{ls}$ from OpenOccupancy~\cite{wang2023openoccupancy}. In addition, we also need a depth supervision loss $L_{d}$ and a 2D semantic loss $L_{s}$, which will be introduced in the next section.

\begin{figure}[htb]
\centering
\includegraphics[width=0.95\linewidth]{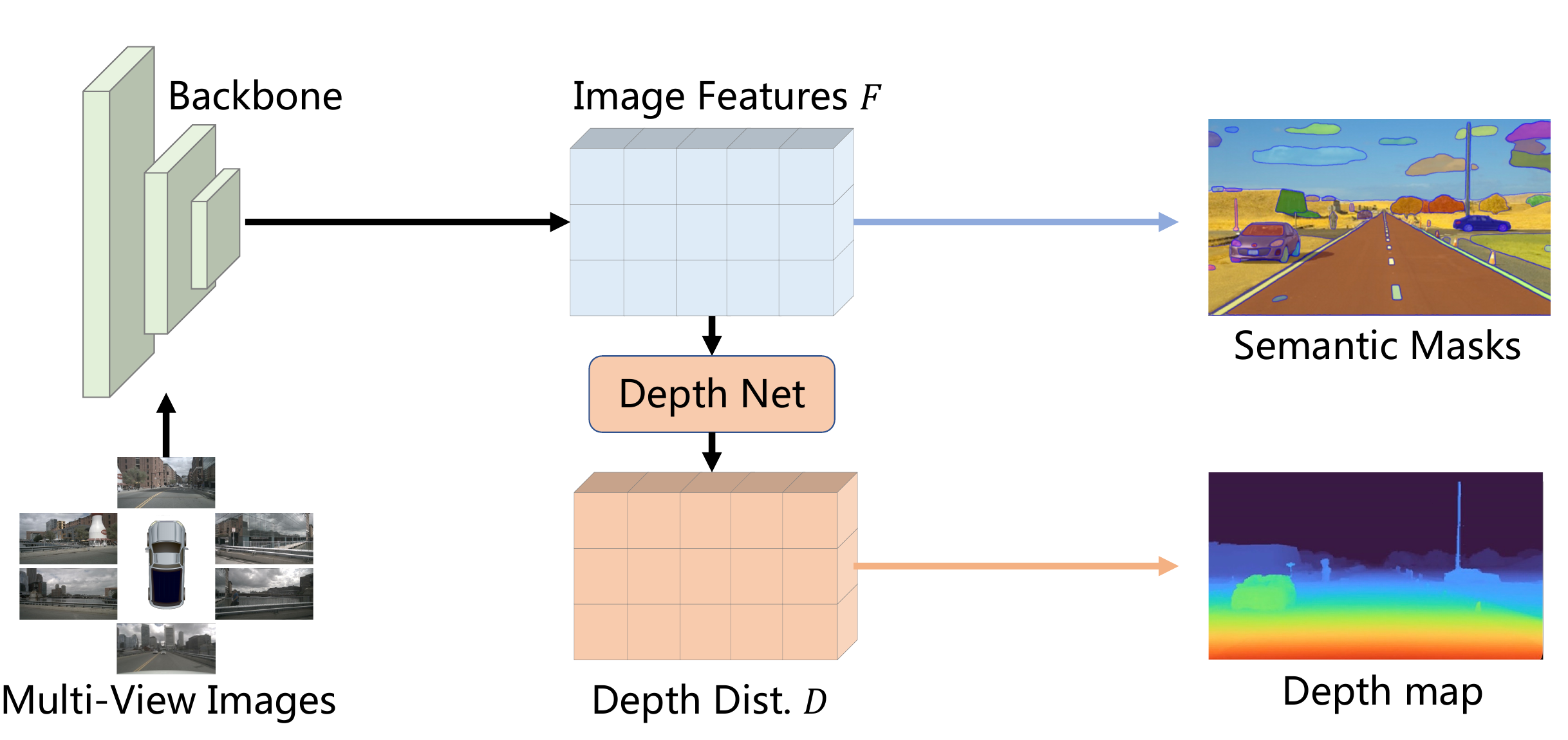}
\caption{\textbf{Joint depth and semantic pre-training.} 
}
\label{fig:decouple}
\vspace{-3mm}
\end{figure}

\subsection{Scaling up and pre-training}\label{SP}

Scaling the model size has traditionally been a convenient approach to improving model accuracy. However, in the field of 3D vision-only perception, researchers have discovered that employing a more powerful 2D backbone often leads to overfitting~\cite{huang2021bevdet}. For instance, on the nuScenes 3D object detection task, the largest backbone, such as VIT-L~\cite{dosovitskiy2021image} with approximately 300M parameters, and commonly used backbones like ConvNext-B~\cite{liu2022convnet} and VoVNet-99~\cite{lee2019energy} with around 100M parameters, tend to encounter this issue.
To address this challenge, we explore the utilization of the 1B-parameter backbone, InternImage-H~\cite{wang2023internimage}, for multi-camera 3D perception tasks. However, directly applying this backbone would result in severe overfitting due to the limited number of samples available for training, specifically the 40K samples in the nuScenes dataset~\cite{caesar2020nuscenes}. To overcome this limitation, we leverage the opportunity provided by this competition, which allows participants to utilize additional public data. By augmenting our data resources, we can train our large-scale model more effectively.
Building upon the open-source InternImage-H checkpoint, we conduct model training on the Object365 dataset~\cite{shao2019objects365}, which is a vast 2D object detection dataset comprising 2 million images. This pre-training on large-scale 2D detection tasks enhances the model's semantic perception capabilities. However, there still exists a certain domain gap when applying the pre-trained model to downstream 3D perception tasks. Therefore, we further perform targeted pre-training on the model specifically for 3D perception tasks.
An effective approach for pre-training is to enhance the model's geometric awareness through depth estimation tasks. Consequently, we conduct extensive pre-training on the nuScenes dataset, primarily focusing on depth estimation. It is worth noting that depth pre-training lacks semantic-level supervision. To mitigate the risk of the model becoming excessively biased towards depth information, potentially leading to the loss of semantic priors (especially given the large-scale nature of the model, which is prone to overfitting), we simultaneously aim to predict the 2D semantic segmentation labels alongside the depth prediction task, as shown in Figure~\ref{fig:decouple}.

However, nuScenes does not provide semantic segmentation labels for 2D images. To address this issue, we employ the popular Segment Anything Model (SAM)~\cite{kirillov2023segment} for automatic labeling. For thing categories with bounding box annotations provided by nuScenes, we utilize box prompts to generate high-quality semantic masks for each object. Unfortunately, for stuff categories such as road surfaces or buildings, bounding box annotations are not available. Nonetheless, nuScenes offers corresponding point cloud semantic segmentation labels for these categories.

To generate semantic masks for the stuff categories, we project LiDAR points belonging to these categories onto the image. For each category, we randomly select three points as point prompts to generate the corresponding semantic masks, which yields satisfactory mask quality. With the 2D image semantic mask labels and the ground truth depth maps, we train the model using a joint depth estimation task and semantic segmentation task. This pre-training task closely aligns with the final occupancy prediction task, enabling the direct generation of 3D occupancy results using depth values and semantic labels. The pre-trained model serves as an improved starting point for the subsequent training of the occupancy prediction task.

\subsection{Post-processing}\label{PP}

\subsubsection{Test-time augmentation}
In test-time augmentation, the input image is horizontally flipped and the 3D space is horizontally and vertically flipped during inference. This gives a total of eight prediction results for each frame. The final prediction is obtained by computing the mean of all the results.

We additionally observe that occupancy prediction accuracy significantly degrades with distance. Temporal test-time augmentation (TTA) is thus used to mitigate this issue. For static voxels, we leverage the predicted voxels that are close to the ego car in previous frames to replace the voxels co-located in the current frame.

\subsubsection{Ensemble}
We perform a weighted sum of all independent results, where the weight of each voxel is determined by two multiplying factors. The first factor is model the weight related to the overall mIoU of each result. The second factor is the specific category weight related to the IoU of this voxel's category. We use NNI~\cite{mnni} to search the different weight values automatically.

\section{Experiments}

\subsection{Datasets and metrics}
\noindent\textbf{Dataset.} The occupancy dataset is built based on the existing nuScenes dataset~\cite{tian2023occ3d,caesar2020nuscenes}. For each frame, they provide occupancy annotations within the range of [-40m, -40m, -1m, 40m, 40m, 5.4m], and the resolution of each voxel is 0.4m. The dataset contains 18 classes, where one indicates a free voxel that is occupied by nothing. The dataset also provides the camera mask to indicate whether the voxel is visible from any cameras.

\vspace{2mm}
\noindent\textbf{Metrics.} For this challenge, we mainly evaluate our models based on mIoU, which can be formulated as follows:
\begin{equation}
    mIoU = \frac{1}{C}\sum_{c=1}^{C}\frac{TP_c}{TP_c+FP_c+FN_c},
\end{equation}
where $TP_c$, $FP_c$, and $FB_c$ correspond to the number of true positive, false positive, and false negative predictions for class $c$, and $C$ is the total number of classes.

\subsection{Implementation details}

\begin{table*}[htb]
\scriptsize
\setlength{\tabcolsep}{0.005\linewidth}
\newcommand{\classfreq}[1]{{~\tiny(\semkitfreq{#1}\%)}}  

\def\mystrut{\rule{0pt}{1.5\normalbaselineskip}}
\centering
\rowcolors[]{2}{black!3}{white}
\begin{adjustbox}{width=1.99\columnwidth,center}
\begin{tabular}{l| c | c c c c c c c c c c c c c c c c c|c}

    \toprule
    Method 
    & \rotatebox{90}{Input}
    & \rotatebox{90}{others} 
    & \rotatebox{90}{barrier}
    & \rotatebox{90}{bicycle} 
    & \rotatebox{90}{bus} 
    & \rotatebox{90}{car} 
    & \rotatebox{90}{construction vehicle} 
    & \rotatebox{90}{motorcycle} 
    & \rotatebox{90}{pedestrian} 
    & \rotatebox{90}{traffic cone} 
    & \rotatebox{90}{trailer} 
    & \rotatebox{90}{truck} 
    & \rotatebox{90}{driveable surface} 
    & \rotatebox{90}{other flat} 
    & \rotatebox{90}{sidewalk} 
    & \rotatebox{90}{terrain} 
    & \rotatebox{90}{manmade} 
    & \rotatebox{90}{vegetation} 
    & \rotatebox{90}{mIoU}  \\
    \midrule
    
    MonoScene~\cite{cao2022monoscene} & C & 1.75 & 7.23 & 4.26 & 4.93 & 9.38 & 5.67 & 3.98 & 3.01 & 5.90 & 4.45 & 7.17 & 14.91 & 6.32 & 7.92 & 7.43 & 1.01 & 7.65 & 6.06 \\
    BEVDet ~\cite{huang2021bevdet} & C & 2.09 & 15.29 & 0.0 & 4.18 & 12.97 & 1.35 & 0.0 & 0.43 & 0.13 & 6.59 & 6.66 & 52.72 & 19.04 & 26.45 & 21.78 & 14.51 & 15.26 & 11.73 \\
    BEVFormer~\cite{li2022bevformer} & C & 5.85 & 37.83 & 17.87 & 40.44 & 42.43 & 7.36 & 23.88 & 21.81 & 20.98 & 22.38 & 30.70 & 55.35 & 28.36 & 36.0 & 28.06 & 20.04 & 17.69 & 26.88 \\
    CTF-Occ~\cite{tian2023occ3d} & C & 8.09 & 39.33 & 20.56 & 38.29 & 42.24 & 16.93 & 24.52 & 22.72 & 21.05 & 22.98 & 31.11 & 53.33 & 33.84 & 37.98 & 33.23 & 20.79 & 18.0 & 28.53 \\
    \midrule
    Version A &C& 0.04&	37.15&	16.81&	34.17&	38.22&	13.41	&16.97&	19.69&	18.94&	11.65&	21.94&	55.94&	26.98&	29.65&	26.92	&10.24	&14.33&	23.12\\ 
    Version B&C & 0.03	&40.94&	21.16&	39.22&	40.75&	20.57	&23.85	&23.6	&24.95&	16.63&	26.36	&59.42	&27.57	&31.39&	29.03&	16.69	&18.42	&27.09\\
    Version C &C& 0.02	&45.18	&25.26	&44.55&	47.38	&22.63	&26.24	&26.92	&27.91&	26.4	&32.1	&76.97	&37.2	&44.84&	47.81	&37.0	&32.64	&35.36
\\
    Version D &C &12.17&	44.83&	25.73&	42.61	&47.97	&23.16	&25.17&	25.77&	26.72	&31.31	&34.89	&78.83&	41.42&	49.06	&52.22	&39.07	&34.61&	37.39
\\
    Version E &C&13.57	&44.74	&27.01	&45.41&	49.1	&25.15	&26.33&	27.86	&27.79&	32.28	&36.75	&80.07&	42.76	&51.18&	55.13	&42.19	&37.53&	39.11
\\Version F&C &
    13.66&	45.88	&28.26&	44.91	&49.78&	26.21&	28.84	&28.27&	27.89	&32.75&	37.56&	81.2	&43.46&	52.13	&56.35&	42.79&	38.1&	39.89
\\Version G & C &
14.41 &45.77 &29.19 &45.29& 50.53&  27.86&  29.01& 28.15& 28.61 &32.89& 37.86&  81.76& 45.52& 53.99 &58.69& 43.49& 38.75& 40.69\\
Version H &C &14.30&49.71&	30.0	&46.62&	51.54	&29.3	&29.13	&29.35	&30.48	&34.97&	39.36	&83.07	&47.16	&55.62	&59.88	&44.89&	39.58	&42.06\\
\bottomrule
\end{tabular}
\end{adjustbox}
\vspace{-2mm}
\caption{3D occupancy prediction performance of different settings on the Occ3D-nuScenes dataset~\cite{tian2023occ3d}.} 
\label{table:occ3d-nus}
\end{table*}

\begin{table*}[htb]
\scriptsize
\setlength{\tabcolsep}{0.005\linewidth}
\newcommand{\classfreq}[1]{{~\tiny(\semkitfreq{#1}\%)}}  

\def\mystrut{\rule{0pt}{1.5\normalbaselineskip}}
\centering
\rowcolors[]{2}{black!3}{white}
\begin{adjustbox}{width=1.99\columnwidth,center}
\begin{tabular}{l| c| c c c c c c c c c c c c c c c c c|c}
    \toprule
    Method 
    & \rotatebox{90}{params.} 
    & \rotatebox{90}{others} 
    & \rotatebox{90}{barrier}
    & \rotatebox{90}{bicycle} 
    & \rotatebox{90}{bus} 
    & \rotatebox{90}{car} 
    & \rotatebox{90}{construction vehicle} 
    & \rotatebox{90}{motorcycle} 
    & \rotatebox{90}{pedestrian} 
    & \rotatebox{90}{traffic cone} 
    & \rotatebox{90}{trailer} 
    & \rotatebox{90}{truck} 
    & \rotatebox{90}{driveable surface} 
    & \rotatebox{90}{other flat} 
    & \rotatebox{90}{sidewalk} 
    & \rotatebox{90}{terrain} 
    & \rotatebox{90}{manmade} 
    & \rotatebox{90}{vegetation} 
    & \rotatebox{90}{mIoU}\\
    \midrule
 Version H  &67.8M&14.30&49.71&	30.0	&46.62&	51.54	&29.3	&29.13	&29.35	&30.48	&34.97&	39.36	&83.07	&47.16	&55.62	&59.88	&44.89&	39.58	&42.06\\
 Version I &130.8M&14.26&57.02&38.34&57.69&62.12&34.35&39.43&38.82&39.42&42.91&50.02&86.04&50.24&60.06&62.54&52.36&45.68&48.90\\
 Version J &428.8M&16.74&55.33&39.77&58.94&61.79&32.04&42.63&40.51&39.06&43.72&51.33&87.34&53.77&62.63&66.06&56.63&49.74&50.47\\
 Version K&  1200.0M & 28.28 & 56.70&44.35& 51.37& 61.81& 35.12&47.38&41.56&39.88 & 57.96&48.39 & 86.66 & 56.97 & 64.66 & 61.23&62.78& 52.35&52.79 \\
\bottomrule
\end{tabular}
\end{adjustbox}
\vspace{-2mm}
\caption{Results of models at different scales.} 
\label{table:main}
\end{table*}

\noindent\textbf{Training Strategies.} For training large-scale models, we use a batch size of 32 on 32 NVIDIA A100 GPUs, AdamW optimizer with a learning rate of $1\times 10^{-4}$ and a weight-decay of 0.05. The learning rate of the backbone is 10 times smaller.
We train our models around 50 epochs for occupancy tasks. 
The temporal windows used by every model are determined based on the GPU memory. 
For the Intern-H backbone, we use 6 previous frames. 
When GPU memory is sufficient, we use up to 16 historical frames.
Following SOLOFusion~\cite{park2023time}, we use  online temporal sequences during training which is much more efficient. 

\vspace{2mm}
\noindent\textbf{Network Details.} For large-scale models, the image features from the backbone are downsampled with a stride of 16. The input image scale is 640$\times$1600. We use commonly used data augmentation strategies, including flip, and rotation on both image and 3D space. The depth net predicts 80 discrete depth categories covering the depth from 2m to 42m. The resolution of generated 3D voxel features is $200\times200\times16$. The backward projection module uses 1 layer since the input BEV queries already contain meaningful information. During the training phase, we ignore the invisible voxels from cameras.

\subsection{Ablation study}

Training large-scale models requires huge computing resources. In our exploration, we first verify the effects of different models at a smaller scale. In this setting, the input scale is $256\times 704$, the resolution is $100\times100\times8$, and the image backbone is ResNet-50~\cite{he2016deep}. We list the milestones of our exploration in Table.~\ref{table:occ3d-nus}.
Version A is our vanilla baseline. In Version B, we use depth supervision following BEVDepth. For Version C, we ignore the invisible voxels from cameras during the training phase. Version D model fixes several major bugs in Version C, especially regarding the abnormal IoU on the \textbf{other} category. For Version E, we use temporal information from the previous 16 frames. We leverage the joint depth and semantic pre-training in Version F. For Version G, we optimize the loss function by adding the Dice loss and using 3D transformation to align voxel features from different timesteps. The results of Version H is the test-time augmentation results of Version G.

\subsection{Scaling up}
After exploring the basic design of FB-OCC, we scale up the model size by using larger backbones and image input size, as shown in Table~\ref{table:main}. Compared to Version H, version I uses VoVNet-99 backbone and other modifications, including using $960\times1760$ image input and a voxel resolution of 0.4m. Compared to Version I, Version J leverages a ViT-L backbone and ViT-Adapter~\cite{chen2023vision}. For our most powerful Version K, we scale up the model to over 1 billion parameters with the InternImage-H backbone.

\subsection{Post-processing}
In our final submission, we adopt our ensemble strategy with seven models for the best accuracy.
The main difference between different models is the use of different backbones. By combining the results of all the models through ensemble techniques, we were able to achieve our best result on the test set with a mIoU score of 54.19\%.

\section{Conclusion}
In this report, we describe our winning solution for the 3D Occupancy Prediction Challenge in conjunction with CVPR 2023. Our solution demonstrates state-of-the-art model design that yields excellent BEV perception. It also shows the effectiveness of visual foundation models and large-scale pre-training in 3D occupancy prediction.

\clearpage
\bibliographystyle{unsrt}
\bibliography{ref}

\end{document}